\documentclass[10pt,twocolumn,letterpaper]{article}

\usepackage[pagenumbers]{cvpr} 

\usepackage{times}
\usepackage{epsfig}
\usepackage{graphicx}
\usepackage{amsmath}
\usepackage{amssymb}
\usepackage{enumitem}
\usepackage{colortbl}
\usepackage{siunitx}
\usepackage{pifont}
\newcommand{\cmark}{\ding{51}}%
\newcommand{\xmark}{\ding{55}}%
\usepackage[accsupp]{axessibility}

\usepackage[pagebackref=true,breaklinks=true,colorlinks,bookmarks=false]{hyperref}

\usepackage{xcolor}
\definecolor{ARcolor}{rgb}{0.0,0.0,0.8}
\definecolor{DMcolor}{rgb}{1.0,0.0,0.0}
\definecolor{KYcolor}{rgb}{0.8,0.8,0.0}

\newcommand{\AR}[1]{\textcolor{ARcolor}{\textbf{AR}:~#1}}
\newcommand{\DM}[1]{\textcolor{DMcolor}{\textbf{DM}:~#1}}

\renewcommand{\paragraph}[1]{\vspace{0.03cm} \noindent{\bf #1}}

\def\cvprPaperID{6607} 
\def\confYear{CVPR 2022}

\usepackage{fancyhdr}
\begin{document}

 \title{Audio-Visual Speech Codecs: \\ Rethinking Audio-Visual Speech Enhancement by Re-Synthesis}

\author{Karren Yang$^1$ \hfill Dejan Markovi\'c$^2$ \hfill Steven Krenn$^2$ \hfill Vasu Agrawal$^2$ \hfill Alexander Richard$^2$\\
$^1$MIT\quad $^2$Meta Reality Labs Research\\
{\tt\small karren@mit.edu \quad \{dejanmarkovic,stevenkrenn,vasuagrawal,richardalex\}@fb.com}
}

\maketitle

\thispagestyle{fancy}

\begin{abstract}
Since facial actions such as lip movements contain significant information about speech content, it is not surprising that audio-visual speech enhancement methods are more accurate than their audio-only counterparts. Yet, state-of-the-art approaches still struggle to generate clean, realistic speech without noise artifacts and unnatural distortions in challenging acoustic environments. In this paper, we propose a novel audio-visual speech enhancement framework for high-fidelity telecommunications in AR/VR. Our approach leverages audio-visual speech cues to generate the codes of a neural speech codec, enabling efficient synthesis of clean, realistic speech from noisy signals. Given the importance of speaker-specific cues in speech, we focus on developing personalized models that work well for individual speakers. We demonstrate the efficacy of our approach on a new audio-visual speech dataset collected in an unconstrained, large vocabulary setting, as well as existing audio-visual datasets, outperforming speech enhancement baselines on both quantitative metrics and human evaluation studies. Please see the supplemental video for qualitative results\footnote{\url{https://github.com/facebookresearch/facestar/releases/download/paper_materials/video.mp4}}.
\vspace{-0.6cm}

\end{abstract}


\section{Introduction}

\begin{figure}[t]
    \centering
    \includegraphics[scale=0.165]{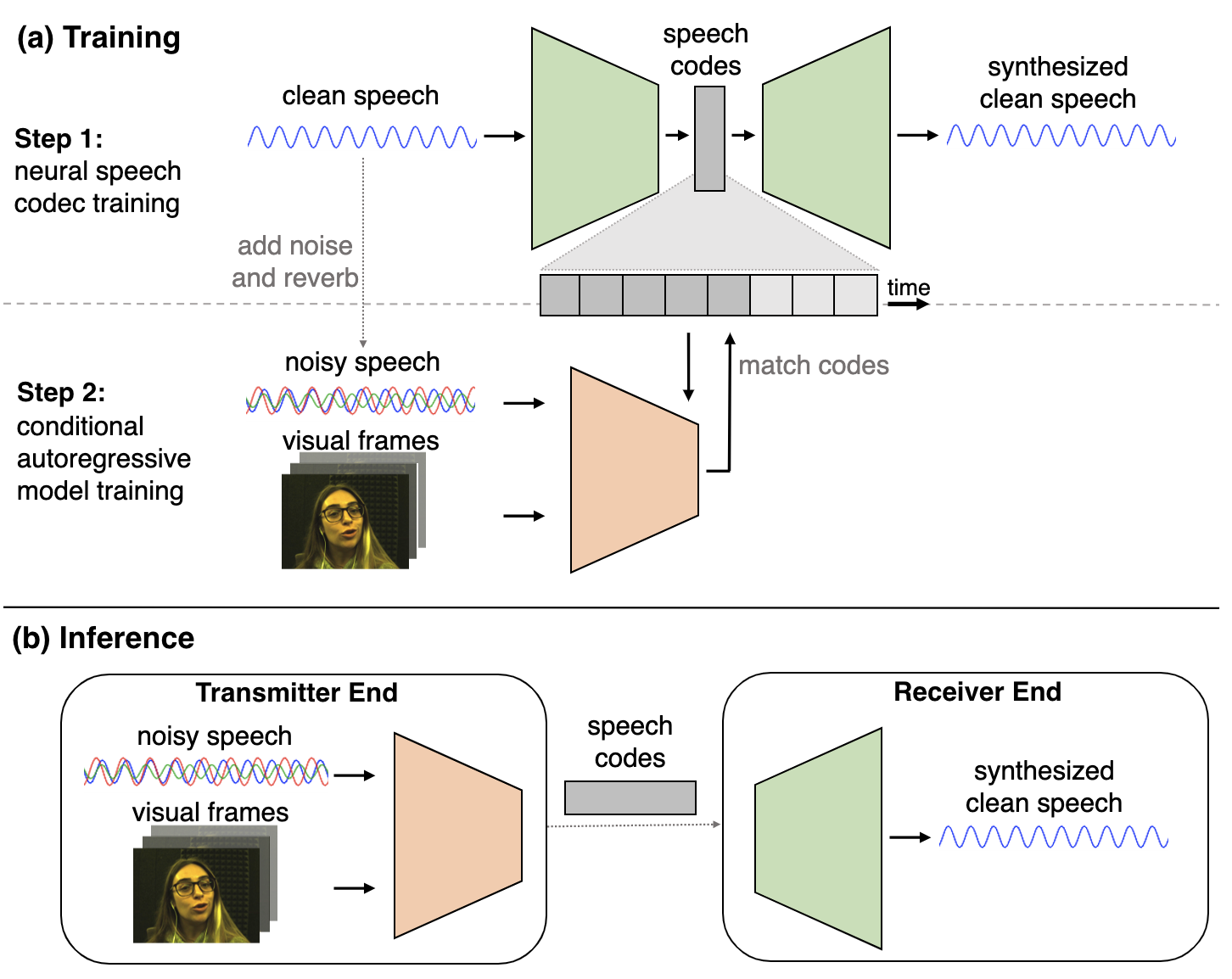}
    \vspace{-0.7cm}
    \caption{\textbf{Audio-Visual Speech Codecs.} Our model performs speech enhancement by leveraging audio-visual speech cues to synthesize the discrete codes of a neural speech codec.
    (a) During training, we first learn a codebook of natural speech for a target speaker by training a neural speech codec to compress and decode their clean speech signal. We then train an auto-regressive probabilistic model over the codes conditioned on noisy audio and visual inputs. 
    (b) During inference, we use the auto-regressive model to generate a sequence of speech codes, which are then synthesized into speech using the decoder module of the speech codec. }
    \label{fig:intro}
    \vspace{-0.5cm}
\end{figure}

Humans have the remarkable ability to extract speech content from visual information such as lip movement. Studies show that viewing speakers' faces improves human listening in noisy environments \cite{sumby1954visual, partan1999communication},
and that individuals naturally learn to read lip movements when their hearing is impaired \cite{iezzoni2004communicating}. Inspired by these observations, audio-visual speech enhancement 
methods leverage the visual input of a speaker to isolate their voice in a noisy environment \cite{michelsanti2021overview}. By integrating facial frames of a target speaker with a noisy audio spectrogram, for example, recent deep learning models can generate a mask for the spectrogram that suppresses irrelevant voices and background sounds from the output \cite{afouras2018conversation, ephrat2018looking, gao2021visualvoice}. These models prove useful for reducing noise and improving speech intelligibility of videos for downstream applications.

However, there are a growing number of telecommunications applications where the quality and realism of the output speech, beyond speech intelligibility, are paramount. One example is social telepresence in AR/VR,
which aims to enable realistic face-to-face conversations between people in a virtual setting \cite{lombardi2018deep, wei2019vr}.
Immersive virtual conversations require extremely high-quality speech signals: each speaker's voice must sound clean and realistic when rendered in the virtual environment, as if a real conversation were taking place there. 
Current state-of-the-art methods fall short of these applications for two main reasons. First, they generate speech by using the noisy audio as a template \cite{afouras2018conversation, ephrat2018looking, gao2021visualvoice} rather than by explicitly modeling the distribution of speech, 
which can lead to bleed-through noise and other unnatural distortions that disrupt the sense of immersion. Second, they focus on learning audio-visual speech cues that generalize well across a large population, but that may fail to capture speaker-specific cues needed for a higher-fidelity model \cite{prajwal2020learning}.

\paragraph{Main Contributions.} In this work, we take a different approach from existing work that overcomes these two limitations. Our main contributions are the following:\\
(1) We propose \textbf{audio-visual (AV) speech codecs}, a novel framework for AV speech enhancement. Rather than using noisy audio input as a template for producing enhanced output, AV speech codecs explicitly model the distribution of speech and re-synthesize clean speech conditioned on audio-visual cues. Our approach is summarized in Figure \ref{fig:intro}. During training, we first learn the building blocks of natural speech by training a neural speech codec to compress and decode clean speech signal through a discrete codebook. Subsequently, we learn an auto-regressive probabilistic model over the codes conditioned on noisy audio and visual inputs. At test time, we obtain speech codes from the auto-regressive model and use the decoder module of the neural speech codec to synthesize clean speech. Our approach is analogous to high-quality, two-stage image generation techniques that learn a probabilistic prior over a pre-trained vocabulary of image components \cite{oord2017neural, esser2021taming}. \\
(2) Rather than adopting a speaker-agnostic framework as done in most recent work, we focus on \emph{personalized models} that leverage speaker-specific audio-visual cues for higher fidelity speech enhancement. To this end, we introduce \textbf{Facestar}%
\footnote{\url{https://github.com/facebookresearch/facestar}}%
, a high-quality audio-visual dataset containing 10 hours of speech data from two speakers. Existing audio-visual datasets used for vision-based speech synthesis tasks are either captured in a clean, controlled environment with a small and constrained vocabulary \cite{cooke2006audio, harte2015tcd} or curated from ``in-the-wild'' videos with variable audio quality and unreliable lip motion \cite{prajwal2020learning}. In contrast, Facestar contains unconstrained, large vocabulary natural speech recorded with high audio and visual quality and enables development of high-quality personalized speech models%
.\\
(3) Empirically, our personalized AV speech codecs outperform audio-visual speech enhancement baselines on quantitative metrics and human evaluation studies while operating on only 2kbps transmission rate from transmitter to receiver. To the best of our knowledge, our work is the first to enable audio-visual speech enhancement at the quality required for high-fidelity telecommunications in AR/VR, even when the transmitter is in a highly noisy and reverberant environment.%

\paragraph{Addressing Scalability.} Beyond introducing high-quality personalized AV speech codecs, we also take steps towards addressing their scalability. While personalized models are commonly used in high-fidelity applications-- for example, personalized visual avatars enable extremely photo-realistic visual representations of humans in VR that overcome the uncanny valley \cite{lombardi2018deep,wei2019vr} -- a downside is that they typically require training on hours of data from the target individual. The question naturally arises of how we can obtain high-quality personalized models with less data, in order to scale high-fidelity telecommunications to a large volume of users. As a first step, we propose a simple strategy for personalizing AV speech codecs to new individuals with minimal new data, based on a similar approach used to scale personalized text-to-speech models \cite{chen2018sample}. Specifically, we introduce a multi-speaker extension of AV speech codecs that features a speaker identity encoder, which can be pre-trained on a multi-speaker dataset and then fine-tuned for a new speaker with only a small sample of their data. We demonstrate this personalization strategy on the GRID dataset \cite{cooke2006audio}. An additional benefit of our multi-speaker model is that it enables voice conversion from one speaker to another, and thereby opens up creative applications in AR/VR.

\section{Related Work}

\paragraph{Audio-only Speech Enhancement.}
The ability of humans to isolate a target speaker from a noisy environment~\cite{sumby1954visual, partan1999communication} has inspired extensive study into computational approaches for speech separation and enhancement. 
While early formulations of the problem assumed input from multiple microphones \cite{yilmaz2004blind, duong2010under}, recent approaches have also considered the monaural setting \cite{roweis2000one, smaragdis2007supervised, spiertz2009source, huang2014deep}. This includes monaural speech separation methods, which address the problem of separating a mixture of speakers from a single audio track \cite{yu2017permutation, huang2014deep, nachmani2020voice, wang2014training}, as well as monaural speech enhancement methods, which tackle the problem of removing non-speech background sounds \cite{weninger2015speech, kumar2016speech, pascual2017segan, stoller2018wave, defossez2020real, su2020hifi} and reverberation \cite{defossez2020real, su2020hifi} from noisy speech. Our work also focuses on the task of monaural speech enhancement, but we diverge from audio-only studies in that we utilize an additional visual stream to guide speech synthesis.

\paragraph{Audio-Visual Source Separation.}
The correspondence between audio and visual cues in video has led to sound source separation approaches that leverage audio-visual information \cite{fisher2001learning, smaragdis2003audio, parekh2017motion, pu2017audio}. Recently, deep learning frameworks have been developed for audio-visual separation for speech \cite{gabbay2017visual, owens2018audio, ephrat2018looking, afouras2018conversation, afouras2019my, chung2020facefilter, gao2021visualvoice} and music \cite{zhao2018sound, zhao2019sound, xu2019recursive, gan2020music}. In the speech domain, these approaches rely on facial recognition \cite{chung2020facefilter, zhao2018sound, gao2021visualvoice} and/or lip motion \cite{owens2018audio, afouras2018conversation, gao2021visualvoice} to suppress sounds that do not correspond to the speaker in the visual stream. We similarly consider the task of audio-visual speech enhancement, but our framework differs from these works in that we perform speech synthesis conditioned on the audio-visual inputs rather than using a sound separation framework (\eg, generating a spectrogram mask). Our results show that our approach leads to higher-quality, more natural sounding speech output.

\paragraph{Neural Speech Codecs.}
Modern audio communication systems rely on speech codecs to efficiently compress and transmit speech. Neural speech codecs use neural networks to compress input speech into a low-bit rate representation that can be transmitted over a network and decoded into audio waveform on the receiver end \cite{vasuki2006review}. The low-bit rate representation is typically a discrete representation of human speech learned through autoencoding \cite{oord2017neural, kankanahalli2018end, garbacea2019low, kleijn2021generative} or speech features obtained through self-supervised training \cite{polyak2021speech}. While some of these approaches consider the impact of noisy speech \cite{kleijn2021generative, zeghidour2021soundstream}, their primary focus is compression of clean speech. In our work, we propose an audio-visual speech enhancement framework that generates the clean speech codes of a neural audio codec conditioned on noisy audio and visual inputs.

\paragraph{Neural Speech Synthesis.}
High-quality neural speech synthesis is generally based on a two-stage pipeline \cite{kong2020hifi}, which first generates a low-resolution intermediate representation of speech from the input and subsequently synthesizes audio waveform from this representation \cite{oord2016wavenet, prenger2019waveglow, kumar2019melgan, kong2020hifi}. 
Prominent examples of speech synthesis include text-to-speech synthesis and video-to-speech synthesis, where the first stage consists of generating the intermediate representation from text \cite{ping2017deep, wang2017tacotron, shen2018natural, li2019neural} or silent video \cite{ephrat2017vid2speech, ephrat2017improved, kumar2019lipper, prajwal2020learning} input. Our approach is similarly based on a two-stage pipeline, but we use learned speech codes from a neural speech codec as our intermediate representation, and we condition the generation of the speech codes on noisy audio and visual inputs, rather than text or silent videos.

\begin{figure*}
    \centering
    \includegraphics[scale=0.12]{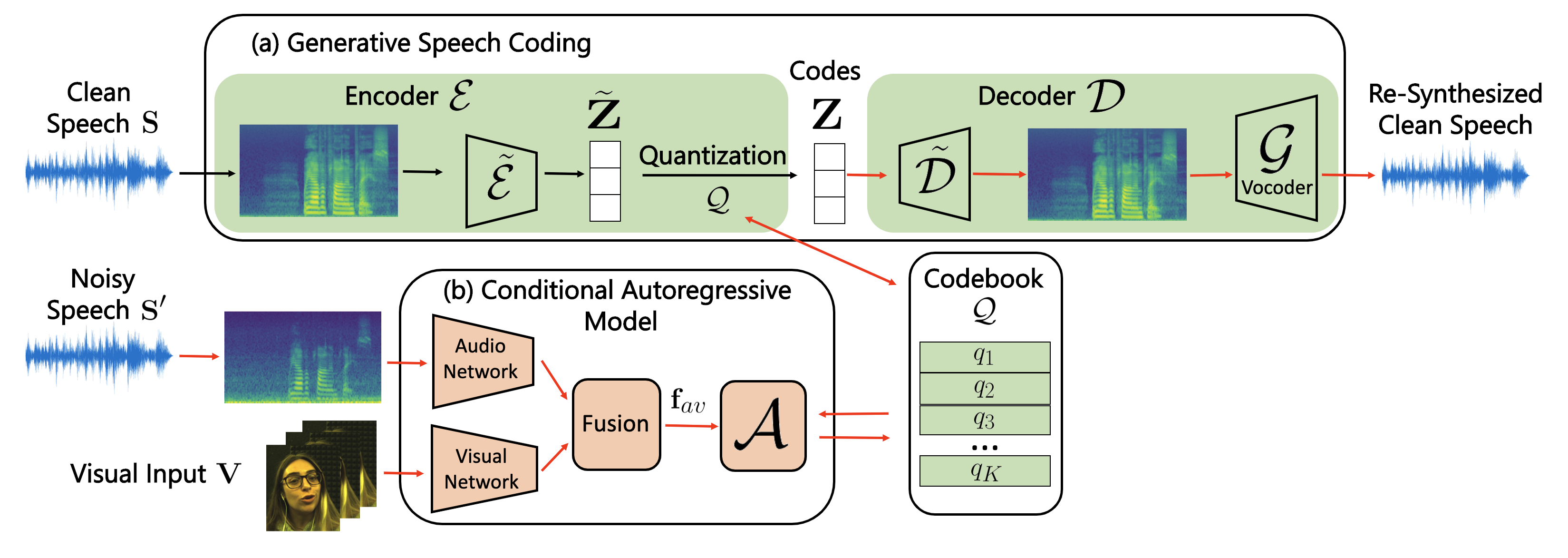}
    \vspace{-0.3cm}
    \caption{\textbf{Schematic of our approach}. (a) We train a Gumbel-Softmax autoencoder on the clean speech of a target speaker to obtain a personalized speech codec. The codec compresses speech into discrete codes that can be re-synthesized into the original speech with high fidelity (Section \ref{sec:generative_speech_coding}). (b) To synthesize clean speech from noisy input, we train an auto-regressive model to generate speech codes conditioned on the noisy audio and lip motion of the target speaker (Section \ref{sec:autoregressive_modeling}). The red arrows indicate information flow at test time. 
    }
    \label{fig:schematic}
    \vspace{-0.3cm}
\end{figure*}

\section{Approach}
Let $\mathbf{S'}$ and $\mathbf{V}$ respectively denote the audio and visual streams of an individual's speech, where $\mathbf{S'}$ contains various sources of environmental noise (\ie, interfering speakers, background sounds, reverberation). Our goal is to synthesize a high-quality, clean version of the speech $\mathbf{S}$. Our approach consists of two learned components, which are summarized in Figure \ref{fig:schematic}: 
\begin{enumerate}
\item \textbf{Generative speech codes}: We learn a discrete speech codebook that captures a rich vocabulary of the target speaker's utterances, which can be used to synthesize high-quality audio in the speaker's voice. %
\item \textbf{Conditional auto-regressive model}: Conditioned on the speaker's audio-visual inputs, we train an auto-regressive model to generate speech codes, ensuring that the sequence of codes follow the natural distribution of the speaker's speech. 
\end{enumerate}
Our framework is inspired by high-quality two-stage generative models, such as  VQ-VAE \cite{oord2017neural}, which first learn a discrete encoding of the data and subsequently learn to generate a code sequence using a probabilistic model. Here, we condition the code generation on the speaker's audio-visual inputs for speech synthesis.

\subsection{Generative Speech Coding}\label{sec:generative_speech_coding}
We represent an individual's clean speech as a sequence of entries from a codebook $\mathcal{Q} = \{q_k\}_{k=1}^K $, where each $ q_k $ is an $N$-dimensional vector. Based on these codes, any clean speech segment $\mathbf{S} \in \mathbb{R}^T$ can be approximately synthesized from a code $\mathbf{Z} \in \mathbb{R}^{T' \times N}$, where $T'$ is the temporal extend of the length $T$ input audio. Since the encoder compresses the input signal along the temporal axis, we typically have $ T' < T $. We first map from speech to codes using the encoder network $\tilde{\mathcal{E}}$, which operates on the mel-spectrogram representation of $\mathbf{S}$:
\begin{equation}
\tilde{\mathbf{Z}} = \tilde{\mathcal{E}}(\textbf{melspec}(\mathbf{S})) \in \mathbb{R}^{T' \times K}.
\end{equation}
Then, we transform this encoding into a sequence of $T'$ codes by sampling from the Gumbel-softmax distribution \cite{jang2016categorical} and selecting the code from $\mathcal{Q}$ with the corresponding index.
For a temporal encoding $ \mathbf{Z} = [\mathbf{Z}_1, \cdots , \mathbf{Z}_{T'}] $, the $t$-th code is therefore given by
\begin{equation}
    \mathbf{Z}_t = q_k, \quad k = \text{Gumbel}(\tilde{\mathbf{Z}}_{t, 1:K}).
\end{equation}
We denote the transformation from continuous valued embeddings $\tilde{\mathbf{Z}}$ to codes $\mathbf{Z}$ by $\mathbf{h}_\mathcal{Q}$. In practice, a codebook that captures the full range of an individual's speech may require a prohibitively large number $K$ of codebook entries.
To increase the expressiveness of the speech codebook, we follow a commonly used concept of \textit{multi-head} codes~\cite{jang2016categorical,richard2021meshtalk} and replace each code in the codebook $\mathcal{Q}$ by a set of $ H $ subcodes
\begin{equation}
    \mathcal{Q}^{(h)} = \{q_k^{(h)}\}_{k=1}^{\tilde K}, \quad h = 1,\dots,H.
\end{equation}
In other words, instead of using one large codebook of size $ K $, we use $ H $ smaller codebooks of size $ \tilde K $ each.
This enables the size of our speech codebook to grow exponentially in $H$, \ie, $K = \tilde K^H$, increasing the expressiveness of our speech codebook without an exponential increase in encoding size. In this case, the dimensionality of the encoding $\tilde{\mathbf{Z}}$ is $T' \times H \times \tilde K$, and each temporal code is
\begin{equation}
    \mathbf{Z}_t = \left[q^{(h)}_k \right]_{h=1}^H, \quad k = \text{Gumbel}(\tilde{\mathbf{Z}}_{t, h, 1:\tilde K}).
\end{equation}

Finally, the decoder that reconstructs speech from the learned codes is composed of a mel-spectrogram decoder $\tilde{\mathcal{D}}$ followed by a neural vocoder $\mathcal{G}$ that transforms the decoded mel-spectrogram back into the wave-domain.
Our speech codec architecture therefore consists of an encoder $ \mathcal{E} = \mathbf{h}_\mathcal{Q} \circ \tilde{\mathcal{E}} $ that maps from mel-spectrograms of the input speech to codes, and a decoder $ \mathcal{D} = \mathcal{G} \circ \tilde{\mathcal{D}} $ that maps from codes back to speech.
The codebook $\mathcal{Q}$ is learned by jointly optimizing it with the encoder $\mathcal{E}$ and the spectrogram decoder $\tilde{\mathcal{D}}$ to minimize the $\ell_2$-loss between the mel-spectrogram of the clean speech signal and the reconstructed mel-spectrogram,
\begin{equation} \label{eq:opt_codebook}
\mathbb{E}_{\mathbf{S}}|| \textbf{melspec}(\mathbf{S}) - \tilde{\mathcal{D}} (\mathbf{Z}) ||_2^2.
\end{equation}

For the neural vocoder, \ie the module that transforms the decoded mel-spectrograms back into a waveform, we use a HiFi-GAN \cite{kong2020hifi} conditioned on the predicted spectrograms $\tilde{\mathcal{D}}(\mathbf{Z})$. The vocoder $\mathcal{G}$ is trained with a combination of a multi-scale GAN loss, a mel-spectrogram loss, and a feature-matching loss as described in Kong \etal \cite{kong2020hifi}. We let $\mathcal{E}$ denote full transformation $\mathbf{S} \mapsto \mathbf{Z}$ and $\mathcal{D}$ denote the full generative speech model $\mathbf{Z} \mapsto \tilde{\mathbf{S}}$.

\subsection{Conditional Auto-Regressive Modeling of Speech Codes}\label{sec:autoregressive_modeling}

Having learned $\mathcal{E}$ and $\mathcal{D}$, any clean speech segment $\mathbf{S}$ can be represented by a sequence of codes $\mathbf{Z}$ that can be used to reconstruct the clean speech $\mathbf{S}$ through $\mathcal{D}$. Therefore, audio-visual speech enhancement can be formulated as an auto-regressive modeling problem in the latent space. Given a sequence of codes $\textbf{Z}_{1:t-1}$, our auto-regressive model $\mathcal{A}$ predicts the distribution of the next codes conditioned on the corresponding audio-visual features $\mathbf{f}^{av}_t$. We use the categorization operator $\mathbf{h}_Q$ to map the log-probability values output by $\mathcal{A}$ to codes in $\mathcal{Q}$, \ie

\begin{equation} \label{eq:ar}
    \mathbf{Z}_t = \mathbf{h}_Q (\mathcal{A}(\mathbf{Z}_{1:t-1}, \mathbf{f}^{av}_t)).
\end{equation}

The audio-visual network that extracts the audio-visual features $\textbf{f}^{av}$ is composed of a visual stream, an audio stream, and an audio-visual fusion module. The visual stream takes $\mathbf{V}$ as input and produces an intermediate representation of visual features of dimension $T' \times H \times \tilde K$ that is useful for shaping speech synthesis. 
%
To produce the audio-visual features $\textbf{f}^{av}$, the visual features are fused with the audio stream along the temporal axis, then passed through a fusion module.
%
The auto-regressive model $\mathcal{A}$ and the audio-visual feature extraction network are optimized to minimize the error in the mel-spectrogram reconstruction,
\begin{equation}\label{eq:opt_ar}
    \mathbb{E}_{(\mathbf{S}, \mathbf{S}', \mathbf{V})} || \textbf{melspec}(\mathbf{S}) - \tilde{\mathcal{D}}(\mathbf{Z})||_2^2.
\end{equation}
where $\mathbf{S}'$ and $\mathbf{V}$ are the noisy input speech and the visual frames, and $\mathbf{Z}$ is the latent code obtained from the conditional auto-regressive model as in Equation \eqref{eq:ar}.

\textbf{Summary: Training vs. Inference.} Training occurs through a two-stage procedure. First, we learn a discrete speech codebook by optimizing Equation (\ref{eq:opt_codebook}) on clean speech data. Any sequence of latent codes can therefore be decoded into (clean) speech using the speech decoder $\mathcal{D}$. Second, we train an auto-regressive model over the codes by optimizing Equation (\ref{eq:opt_ar}) using noisy audio and visual data. Note that the speech codebook and decoder are fixed during this step. At inference time, information flow follows the red arrows in Figure \ref{fig:schematic}. We first predict speech codes $\mathbf{Z}$ in an auto-regressive fashion through Equation \eqref{eq:ar}. Then, the codes are passed through the speech decoder $\mathcal{D}$ to re-synthesize the output speech. Note the advantage of this two-step approach: since the decoder is trained on clean speech only and the speech codebook has limited capacity, the decoder is incapable of producing non-speech-like outputs. In contrast to existing approaches, bleeding-through of noise from the noisy input speech can therefore be avoided entirely.

\section{Facestar Dataset} \label{sec:facestar}

Existing audio-visual datasets tend to be either (i) captured in a clean, controlled environment with a small and constrained vocabulary such as GRID \cite{cooke2006audio} or TCD-TIMIT \cite{harte2015tcd}; or (ii) curated from ``in-the-wild'' videos with variable audio quality and unreliable lip motion such as Lip2Wav \cite{prajwal2020learning}. The former datasets are constrained and do not cover the range of natural speaker content, while the latter does not have sufficiently high audio quality for training high-quality speech synthesis models.
The necessity for a dataset with high-quality audio and video in a conversational, large-vocabulary setting becomes apparent for applications such as video calls, where background noise poses a significant limitation to a socially engaging experience.
Therefore, we collect and introduce the Facestar dataset, which consists of 10+ hours of audio-visual speech data collected in a video-conferencing setting. The dataset was recorded in a low-noise acoustically treated 2.5\si{\cubic\metre} environment with pseudo uniform lighting.
Two participants, one male and one female, spoke freely in front of a video-conferencing device equipped with visual and audio sensors. For technical details see supplemental material. Each participant was captured for 5+ hours, resulting in 10+ hours of high-quality audio-visual data
containing frontal face view and natural unconstrained speech that simulates typical video call settings.
Table \ref{table:dataset} shows a comparison of the Facestar dataset with existing single-speaker datasets used for audio-visual tasks. Compared to other datasets, the Facestar dataset contains unconstrained speech captured in a clean, video-conferencing like environment, providing the type of high-quality audio needed for training speech synthesis models. 

\begin{table}[t]
\scriptsize
\centering
\begin{tabular}{|c|c|c|c|c|}
\hline \rowcolor[gray]{0.8}
\textbf{\begin{tabular}[c]{@{}c@{}}AV \\ Dataset\end{tabular}} & \textbf{\begin{tabular}[c]{@{}c@{}}\# Hours \\ per \\ Speaker\end{tabular}} & \textbf{\begin{tabular}[c]{@{}c@{}}High-\\ Quality\\ Audio\end{tabular}} & \textbf{\begin{tabular}[c]{@{}c@{}}Reliable\\ Lip\\ Motion\end{tabular}} & \textbf{\begin{tabular}[c]{@{}c@{}}Unconstrained\\ Natural \\ Speech \end{tabular}} \\ \hline \rowcolor[gray]{0.99}
\textbf{GRID}                                                  & 0.8                                                                         & \cmark                                                                        & \cmark                                                                        & \xmark                                                                       \\  \rowcolor[gray]{0.9}
\textbf{TCD-Timit}                                             & 0.5                                                                         & \cmark                                                                        & \cmark                                                                        & \xmark                                                                       \\  \rowcolor[gray]{0.99}
\textbf{Lip2Wav}                                               & 20                                                                          & \xmark                                                                        & \xmark                                                                        & \cmark                                                                       \\  \rowcolor[gray]{0.9}
\textbf{Facestar}                                              & 5                                                                     & \cmark                                                                        & \cmark                                                                        & \cmark                                                                       \\ \hline
\end{tabular}
\vspace{-0.3cm}
\caption{\textbf{Comparison of different audio-visual speech synthesis datasets}. Our Facestar dataset contains significantly more content per speaker compared to GRID \cite{cooke2006audio} and TCD-Timit \cite{harte2015tcd} and contains higher-quality data compared to the YouTube-Lip2Wav  \cite{prajwal2020learning}.}
\label{table:dataset}
\vspace{-0.3cm}
\end{table}

\section{Experiments on Single-Speaker Datasets}

\begin{table*}[t]
\scriptsize
\centering
\begin{tabular}{|l|rrrrr|rrrrr|}
\hline \rowcolor[gray]{0.8}
               & \multicolumn{5}{c|}{\textbf{Facestar}}    &   \multicolumn{5}{c|}{\textbf{Lip2Wav}} \\ \rowcolor[gray]{0.8}
\textbf{Model} & \textbf{PESQ $\uparrow$} & \textbf{STOI $\uparrow$} & \textbf{F-SNR $\uparrow$} & \textbf{MCD $\downarrow$} & \textbf{Mel-$\ell_2$ $\downarrow$}     &    \textbf{PESQ $\uparrow$} & \textbf{STOI $\uparrow$}     & \textbf{F-SNR $\uparrow$}      & \textbf{MCD $\downarrow$}      & \textbf{Mel-$\ell_2$ $\downarrow$} \\ \hline \rowcolor[gray]{0.99}
Demucs \cite{defossez2020real} & 1.251 & 0.554 & 5.602 & 5.003 & 0.0106      &      1.383     & 0.672       & 7.644       & 4.724       & 0.0109\\  \rowcolor[gray]{0.9}
AV-Masking \cite{gao2021visualvoice} & 1.257 & 0.593 & 5.991 & 5.184 & 0.0093     &      1.438     & 0.689       & 7.873       & 5.167       & 0.0093\\  \rowcolor[gray]{0.99}
AV-Mapping \cite{gabbay2017visual} & 1.332 & 0.626 & 2.802 & 4.885 & 0.0059      &      1.417     & 0.661       & 6.892       & 4.643       & 0.0062\\ \rowcolor[gray]{0.9}
Ours & \textbf{1.354} & \textbf{0.661} & \textbf{7.322} & \textbf{3.815} & \textbf{0.0056} &  \textbf{1.482} & \textbf{0.740}  & \textbf{8.801}  & \textbf{4.072}  & \textbf{0.0055}  \\ \hline \rowcolor[gray]{0.9}
\end{tabular}
\vspace{-0.3cm}
\caption{\textbf{Quantitative Evaluation of Audio-Visual Speech Separation and Enhancement.} Our approach consistently outperforms the baselines on both datasets. For PESQ, STOI, F-SNR, higher is better. For MCD and Mel-$\ell_2$, lower is better.}
\label{table:quantitative_comparison}
\vspace{-0.3cm}
\end{table*}

In this section, we train and evaluate personalized models on large, single-speaker datasets and demonstrate significant quantitative and perceptual gains over baseline methods trained on the same data. We introduce a multi-speaker extension that addresses scalability later in Section~\ref{sec:scalability_to_multispeaker}.

\subsection{Evaluation Setup}

\paragraph{Datasets.} We train and evaluate our approach on the \textit{Facestar} dataset described in Section \ref{sec:facestar}.
Additionally, we evaluate our approach on the \textit{Lip2Wav} dataset \cite{prajwal2020learning}. This dataset consists of videos downloaded from YouTube channels with approximately 20 hours of speech available per speaker.


\paragraph{Model Architecture.}\label{sec:architecture}
We describe the architecture of the generative speech coding model (Figure~\ref{fig:schematic}a) and the conditional auto-regressive model (Figure~\ref{fig:schematic}b).

\paragraph{(a) Generative Speech Coding.} Our encoder network $\mathcal{E}$ consists of a 1D convolutional block (512 filters, kernel size=5, stride=1) with batch normalization and ReLU activation, followed by three 1D residual blocks with the same hyperparameters. The projection to $\tilde{\mathbf{Z}}$ is performed by a final 1D convolutional layer (kernel size=1, stride=1) where the filter size depends on the size of the discrete latent space. For our discrete latent space, we use $\tilde{K}=256, H=4, N=64$. The decoder network $\mathcal{D}$ resembles the encoder network, except there are three additional LSTM modules after the three residual blocks. For our neural vocoder $\mathcal{G}$, we use the architecture of HiFi-GAN \cite{kong2020hifi}.

\paragraph{(b) Conditional Auto-Regressive Model.} Our audio processing network consists of a 1D convolutional block (512 filters, kernel size=5, stride=1) with batch normalization and ReLU activation, followed by three 1D residual blocks with the same hyperparameters. Our visual processing network consists of a 3D convolutional block (64 filters, kernel size=(5,7,7), stride=(1,2,2)) with batch normalization and ReLU activation, followed by a max pooling layer (kernel size=(1,3,3), stride=(1,2,2)). The resulting 4D tensor is passed through the feature extractor of a 2D ResNet \cite{he2016deep}, which acts independently on each temporal frame.
The output
is upsampled to match the temporal frequency of the processed audio and passed through a series of 1D convolutional and residual blocks, similar to the audio processing network. Finally, the audio and visual features are fused by matching their temporal axes and passed through an autoregressive module, which consists of two LSTMs with a PreNet (2 fully-connected layers with dropout) for processing previous frames \cite{shen2018natural}, and a fully-connected layer for mapping the output of the LSTMs to codes.

\paragraph{Training Details.} \label{sec:training}
We train our model on 3-second clips of clean speech randomly sampled from the target dataset, which corresponds to 75-90 video frames. The video frames are pre-processed using the $\text{S}^3\text{FD}$ face detector \cite{zhang2017s3fd} to obtain face crops as done in Prajwal \etal \cite{prajwal2020learning}. To simulate noisy speech at training time, we first add room reverberation by convolving clean speech with impulse responses from the MIT Impulse Response Survey \cite{traer2016statistics}. The MIT Impulse Response Survey consists of 270 impulse responses collected from different locations that volunteers encountered in their daily lives and reflect a variety of standard acoustic settings. Subsequently, we add random audio samples from the VoxCeleb2 \cite{chung2018voxceleb2} dataset to simulate interfering speakers, and random audio samples from Audioset \cite{gemmeke2017audio} to simulate background noise. For the background noise, we adjust the signal-to-noise ratio randomly between 0 and 40 dB with respect to the clean signal.

\subsection{Baselines}
State-of-the-art deep learning-based AV speech enhancement approaches generally fall into two camps: \emph{spectrogram masking} approaches that leverage the visual modality to mask the noisy audio spectrogram \cite{gao20192, gao2021visualvoice}, and \emph{direct mapping} approaches that leverage the audio and visual modalities to directly generate a denoised spectrogram \cite{gabbay2017visual}. We compare our approach to a model from each camp:

\paragraph{Audio-Visual Spectrogram Masking (AV-Masking).} For the spectrogram masking baseline, we use a U-Net model that integrates visual information in the bottleneck layer and outputs a mask for a complex spectrogram \cite{gao20192, gao2021visualvoice}. The mask is applied to the complex spectrogram of the input audio and the result is transformed back into time domain using an inverse FFT to generate the output waveform.

\paragraph{Audio-Visual Direct Mapping (AV-Mapping).} For the direct mapping baseline, we use an encoder-decoder architecture that directly outputs a denoised spectrogram from audio-visual input and uses the Griffin-Lim algorithm to generate the output waveform \cite{gabbay2017visual}. 

In addition to these audio-visual baselines, we also compare against a well-known \textbf{Audio-Only Model} \textbf{(Demucs)} \cite{defossez2020real} to demonstrate the importance of the visual stream for the speech enhancement task. All models are trained and evaluated on the same datasets for fair comparison.





\subsection{Quantitative Evaluation} \label{sec:quantitative_results}

\paragraph{Metrics.} 
We evaluate all approaches using the following metrics: Perceptual Evaluation of Speech Quality, which measures speech quality (\textbf{PESQ}, higher is better); Short-Time Objective Intelligibility, which measures speech intelligibility (\textbf{STOI}, higher is better); Frequency-Weighted Segmental Signal-to-Noise Ratio (\textbf{F-SNR}, higher is better); and the $\ell_2$ distance between the mel-frequency cepstrum coefficients and mel-spectrograms of predicted and ground truth audio (\textbf{MCD}, \textbf{Mel-$\ell_2$}, lower is better). 

\paragraph{Results.} The results on Facestar and Lip2Wave are shown in Table \ref{table:quantitative_comparison}.
Our approach outperforms the baselines on all of the objective metrics used to evaluate enhanced speech. The audio-visual models (AV-Masking, AV-Mapping, ours) generally outperform the audio-only model (Demucs), demonstrating the importance of leveraging information from the visual modality regardless of the specific speech enhancement framework.



\subsection{Human Evaluation Studies} \label{sec:user_study_results}

Although the objective metrics shown in Table \ref{table:quantitative_comparison} are widely used in literature, it is important to note that no objective metric precisely reflects how humans perceive speech quality \cite{kinoshita2016summary}. Therefore, we also conduct a user study using the Facestar dataset to compare our model results to the two audio-visual baselines as well as ground truth recordings. In the study, participants were presented two clips of the same sequence from two different approaches. 
The clips were presented in random order to ensure an unbiased evaluation, and participants were asked to decide which of the two clips sounded more natural, with three answer options: first clip, second clip, or can not tell. Overall, 100 participants ranked around 1000 clips. 
The results in Table \ref{table:perceptual_evaluation} show that our approach is strongly preferred over the baselines. Notably, over 50\% of the time, study participants could not differentiate the outputs of our model from the ground truth recording, which indicates the high quality of our personalized approach.




\begin{table}[]
\scriptsize
\centering
\begin{tabular}{|c|c|c|}
\hline
\rowcolor[gray]{0.8}
\textbf{\begin{tabular}[c]{@{}c@{}}Ours\end{tabular}}                                                      & \textbf{\begin{tabular}[c]{@{}c@{}}GT recordings\end{tabular}}            & \textbf{\begin{tabular}[c]{@{}c@{}}Can not tell\end{tabular}}                    \\ \hline
\rowcolor[gray]{0.99}
4.1\% & 44.5\%          & 51.4\%                    \\ \hline \hline
\rowcolor[gray]{0.8}
\textbf{\begin{tabular}[c]{@{}c@{}}Ours \end{tabular}}                                                      & \textbf{\begin{tabular}[c]{@{}c@{}}AV Encoder Decoder\end{tabular}}            & \textbf{\begin{tabular}[c]{@{}c@{}}Can not tell\end{tabular}}                     \\ \hline
\rowcolor[gray]{0.99}
73.3\% & 6.0\% & 20.7\%                    \\ \hline \hline
\rowcolor[gray]{0.8}
\textbf{\begin{tabular}[c]{@{}c@{}}Ours \end{tabular}}                                                      & \textbf{\begin{tabular}[c]{@{}c@{}}AV Masking \end{tabular}}            & \textbf{\begin{tabular}[c]{@{}c@{}}Can not tell\end{tabular}}    \\ \hline
\rowcolor[gray]{0.99}
78.5\% & 5.7\% & 15.8\%                    \\ \hline 
\end{tabular}
\vspace{-0.3cm}
\caption{\textbf{Perceptual Evaluation}. Participants were presented two  video clips and asked to tell which of the two sounds more natural.}
\label{table:perceptual_evaluation}
\vspace{-0.2cm}
\end{table}

\begin{table}[]
\scriptsize
\centering
\begin{tabular}{|l|cc|}
\hline \rowcolor[gray]{0.8}
& \multicolumn{1}{c}{\textbf{reverb + noise}} & \multicolumn{1}{c|}{\textbf{only}} \\ \rowcolor[gray]{0.8}
\textbf{Model} & \multicolumn{1}{c}{\textbf{+ interfering spkr}} & \multicolumn{1}{c|}{\textbf{reverb + noise}} \\ \hline \rowcolor[gray]{0.99}
\textbf{Vision-Only} & \multicolumn{2}{c|}{0.0085} \\  \rowcolor[gray]{0.9}
\textbf{Audio-Only} & 0.0091 & 0.0056 \\  \rowcolor[gray]{0.99}
\textbf{No Auto-Regressive Module} & 0.0051 & 0.0036 \\  \rowcolor[gray]{0.9}
\textbf{Full Model} & \textbf{0.0043} & \textbf{0.0033}\\ \hline \rowcolor[gray]{0.99}
\end{tabular}
\vspace{-0.3cm}
\caption{\textbf{Ablation Results}. The values shown are the mean $\ell_2$ errors between predicted and ground truth mel-spectrograms for ablation models trained on the Facestar dataset (Speaker 1); lower is better. See text for details.}
\label{table:ablation}
\vspace{-0.5cm}
\end{table}

\subsection{Ablation Studies}\label{sec:ablation}

\begin{figure}
    \centering
    \includegraphics[scale=0.27]{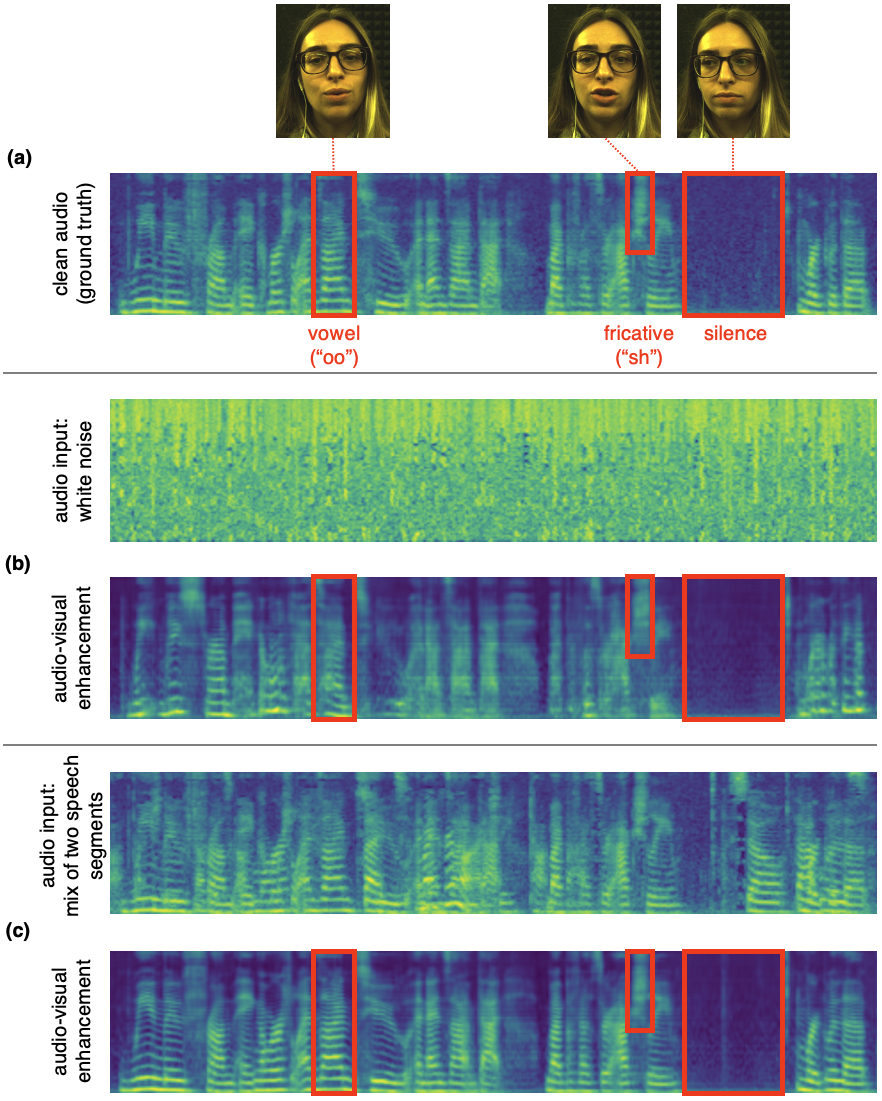}
    \vspace{-0.5cm}
    \caption{\textbf{Importance of visual modality.} (a) Ground truth mel-spectrogram and frames from visual input corresponding to specific vocal sounds. (b) Presented with white noise as audio input, the model relies on the visual modality to synthesize speech. (c) Presented with a speech mixture from the same speaker as audio input, the model relies on the visual modality to disambiguate and separate speech signals. See supplemental video for examples.}
    \label{fig:av_comparison}
    \vspace{-0.5cm}
\end{figure}

\paragraph{Importance of Visual Modality.} Since the noisy audio input contains significant corruptions (\ie, overlapping speakers, high-intensity noise, reverberation), the visual modality is key to synthesizing the target speech components. Table~\ref{table:ablation} shows ablation results for a vision-only model (row 1) and audio-only model (row 2) compared to the full audio-visual model (row 4). Note that model performance declines significantly without the visual modality (compare row 4 to row 2). The visual modality also plays a larger role when the noisy audio contains interfering speakers (column 1) compared to when it contains only background noise and reverberation (column 2), as visual information is needed to disambiguate between speakers. To further support this point, in the presence of interfering speakers, we find that the visual-only model outperforms the audio-only model (compare row 1 to row 2, column 1).

Figure \ref{fig:av_comparison} illustrates how the visual modality is used by our model. The visual input of the target speaker contains specific mouth articulations that correspond to sounds (\eg, vowels and fricatives) in the ground truth speech (Figure \ref{fig:av_comparison}a). Our model is able to approximate these sounds from the visual cues alone, when only white noise is provided to the audio stream (Figure \ref{fig:av_comparison}b). This suggests that the visual modality is primarily responsible for defining the structure of the synthesized speech from our model. When presented with noisy audio input, our model is able to further refine the pitch of the synthesized speech, since this information is not available from visual cues (Figure \ref{fig:av_comparison}c). Note that the noisy audio example in Figure \ref{fig:av_comparison}c contains interfering speech from the same speaker, \ie real and interfering speech share the \textit{same voice} and the visual modality is required in order to disambiguate between target and interfering speech signals. The model successfully leverages the visual cues to suppress the distractor speech and only keeps the original speech signal.


\paragraph{Importance of Auto-Regressive Modeling.} Our conditional auto-regressive model generates speech codes conditioned on previous speech codes, ensuring that the sequence of speech codes is temporally consistent. To determine the contribution of this model component, we perform an ablation study using a model without the auto-regressive component. As shown in Table \ref{table:ablation} (compare row 3 to row 4), this leads to a significant decrease in model performance. 

\paragraph{Efficiency.}
Discretized neural speech codecs generally allow for highly efficient signal transmission.
For instance, Soundstream~\cite{zeghidour2021soundstream} demonstrates compelling speech reconstruction with bitrates from 6-12kbps.
In contrast, our approach is a personalized model and therefore allows for an even stronger data compression with no or barely noticeable loss of quality.
Results shown in the supplemental video have a bitrate of 2kbps, which we found to be sufficient for personalized speech reconstruction.

\begin{figure}
    \centering
    \includegraphics[scale=0.1]{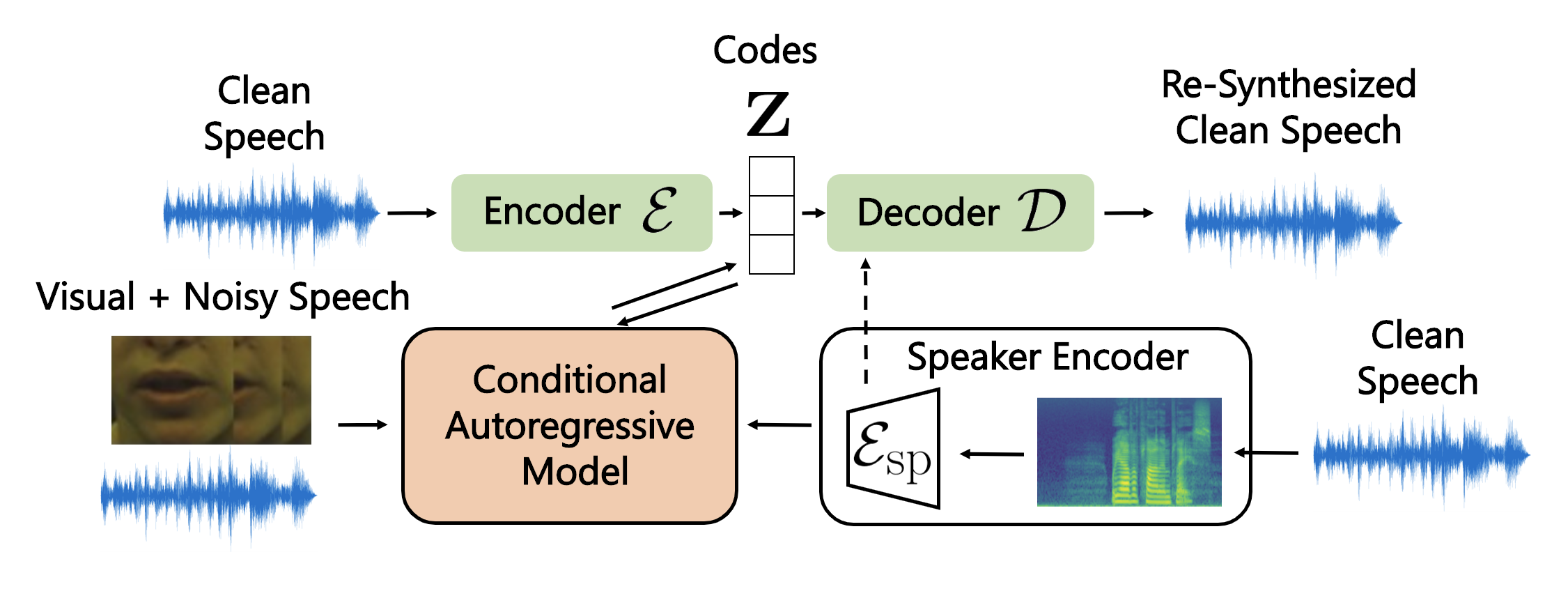}
    \vspace{-0.5cm}
    \caption{\textbf{Multi-speaker model}. A speaker encoder is added to the pipeline from Figure~\ref{fig:schematic}. Restricting the size of the codebook forces the model to disentangle speech content and speaker identity as shown in~\cite{qian2019autovc}.}
    \label{fig:schematic_multispeaker}
\end{figure}

\begin{table}[]
\centering
\scriptsize
\begin{tabular}{|l|rrrr|}
\hline \rowcolor[gray]{0.8}
\textbf{} & \multicolumn{4}{c|}{\textbf{GRID Speaker}} \\ \rowcolor[gray]{0.8}
\textbf{} & \textbf{Sp.~1 (M)} & \textbf{Sp.~3 (M)} & \textbf{Sp.~11 (F)} & \textbf{Sp.~15 (F)} \\ \hline \rowcolor[gray]{0.99}
\textbf{Single-speaker model} & 0.00509 & 0.00794 & 0.00746 & 0.00781 \\  \rowcolor[gray]{0.9}
\textbf{Multi-speaker model} & 0.00657 & 0.00909 & 0.00960 & 0.01594 \\  \rowcolor[gray]{0.8}
\hline
\multicolumn{5}{|c|}{\textbf{Multi-speaker model personalized to new speaker with $k$ minutes of data}} \\ \rowcolor[gray]{0.99}
\hline
\textbf{5 min} & 0.00481 & 0.00682 & 0.00625 & 0.00681 \\  \rowcolor[gray]{0.9}
\textbf{12.5 min} & 0.00457 & 0.00620 & 0.00589 & 0.00655 \\  \rowcolor[gray]{0.99}
\textbf{25 min} & 0.00443 & 0.00595 & 0.00570 & 0.00621 \\  \rowcolor[gray]{0.9}
\textbf{50 min} & 0.00425 & 0.00561 & 0.00553 & 0.00596 \\ \hline 
\end{tabular}
\caption{\textbf{Performance of multi-speaker models that are personalized to new speakers by fine-tuning on different quantities of target speaker data}. The personalized (\ie, fine-tuned) models outperform the single-speaker models even when the amount of data for fine-tuning is greatly reduced. Values shown represent mean $\ell_2$ distances between predicted and ground truth mel-spectrograms; lower is better. }
\label{table:multi_vs_single}
\vspace{-0.3cm}
\end{table}

\begin{figure*}[t]
    \centering
    \includegraphics[scale=0.4]{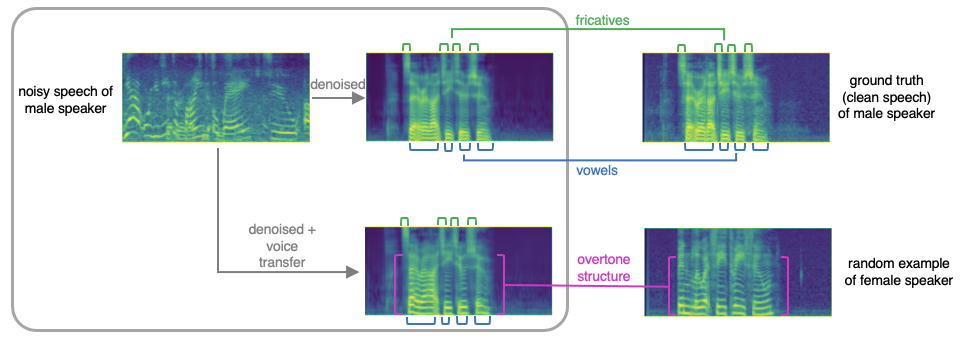}
    \vspace{-0.3cm}
    \caption{\textbf{Voice Transfer Examples.} By swapping the speaker code at the decoder stage, we can synthesize clean audio in a different target speaker's voice. Images shown are mel-spectrogram representations of audio. Note how the linguistic content (\ie, vowels and fricatives) are carried over from the original male speaker, while the pitch and overtone structure are changed to that of the female speaker.}
    \label{fig:voice_transfer}
    \vspace{-0.5cm}
\end{figure*}

\section{Scalability with Multi-Speaker Extension}
\label{sec:scalability_to_multispeaker}

So far we have demonstrated the efficacy of AV speech codecs as a personalized model when they are trained on single-speaker datasets comprised of hours of data. In real-world applications involving a high volume of telepresence users, however, one must be able to obtain high-quality personalized models with less individualized data. In this section, we extend our work to the multi-speaker setting, demonstrating two advantages to such a model: (i) efficient model personalization and (ii) voice controllability.

\paragraph{Efficient Model Personalization.} For practical applicability, there is a need for personalized speech models that can synthesize speech for new users given a small sample of audio-visual data. We extend our framework to the multi-speaker setting by adding a speaker identity encoder to the model as shown in Figure \ref{fig:schematic_multispeaker}, following the approach of personalized text-to-speech synthesis models \cite{chen2018sample}. Our multi-speaker AV speech codec can be pretrained on a larger, multi-speaker dataset and finetuned on a small amount of a new speaker's data to personalize the model to their speech. Table \ref{table:multi_vs_single} shows the results of pretraining on 30 speakers from the GRID dataset \cite{cooke2006audio} and finetuning on different amounts of data for four held-out speakers (last four rows). For comparison, we train four single-speaker models on all data for the held-out speakers (row 1). We find that personalizing our multi-speaker model to new speakers significantly reduces the amount of data needed to achieve the same performance of a single-speaker model:
with only 5 minutes of data per speaker, the fine-tuned multi-speaker model already outperforms the pure single-speaker models.
Note that this performance gain is in fact attributed to the fine-tuning process; the multi-speaker model alone does not generalize well to held-out speakers and is therefore strictly worse than single-speaker models.

\paragraph{Voice Controllability.} An additional benefit of the multi-speaker extension is the ability to transfer from the source speaker's voice to a different target speaker's voice. During training of the speech codec, adding a path for information flow from the speaker identity encoder directly to the decoder as shown in Figure \ref{fig:schematic_multispeaker} and restricting the size of the codebook disentangles speech content from speaker identity; the information bottleneck forces speech codes to reflect speech content \cite{qian2019autovc}. Voice transfer is therefore achieved by swapping the speaker identity embedding with the identity embedding of the new target speaker.
Figure~\ref{fig:voice_transfer} shows an example of such a voice transfer.
Denoising alone restores the linguistic content, \eg the vowels and fricatives, of the noisy input speech.
Denoising with simultaneous voice transfer (by swapping the speaker embedding to another speaker) produces a result in which the same speech content (\eg vowel and fricatives in Figure~\ref{fig:voice_transfer}) are maintained but additionally the overtone structure, which determines the sound of one's voice, is adjusted according to the new speaker identity.

\vspace{-0.3cm}

\section{Conclusion}
We presented a novel speech enhancement framework that maps video and noisy audio inputs onto discrete speech codes, from which clean speech can be re-synthesized without bleeding-through of acoustic noise or unnatural distortions. To train and evaluate our model, we introduced a novel audio-visual dataset containing more than 10 hours of unconstrained, natural speech with large-vocabulary and high-quality audio and visual recordings. Experiments show that our approach outperforms existing frameworks both in quantitative evaluation and human perceptual studies.
In the same way that personalized photo-realistic codec avatars are pushing 3D face representations beyond the uncanny valley, we show that personalized audio-visual speech codecs enable a similar leap forward in audio-visual speech enhancement for VR telepresence applications.


\paragraph{Limitations.} (1) Personalized AV speech codecs 
require a separate model for each user, which comes at a higher computational cost than speaker-agnostic methods. We propose a first step towards scaling up in Section \ref{sec:scalability_to_multispeaker}, but a large-vocabulary multi-speaker dataset with high-quality audio is needed to investigate this direction further. (2) When our model fails (\eg, in very noisy settings), the outputs may be realistic but do not faithfully represent the user's speech. 
In extreme cases, our model can hallucinate plausible mumbling that was not part of the user's original speech.


\paragraph{Ethical Considerations.} As with all speech synthesis systems that enable voice conversion, our approach has to be handled responsibly to avoid audio deep-fakes. Audio watermarking \cite{arnold2000audio} is one strategy for protecting against misuse.




{\small
\bibliographystyle{ieee_fullname}
\bibliography{egbib}
}

\end{document}